# Mixture separability loss in a deep convolutional network for image classification

Trung Dung Do[1*], Cheng-Bin Jin[1†], Van Huan Nguyen[2‡], Hakil Kim[1§]

[1]Department of Information and Communication Engineering, Inha University, 22212, Incheon, Korea
[2]Faculty of Information Technology, Ton Duc Thang University, Ho Chi Minh City, Vietnam
*dotd@inha.edu, †chengbinjin@inha.edu, ‡nguyenvanhuan@tdt.edu.vn, §hikim@inha.ac.kr

**Abstract:** In machine learning, the cost function is crucial because it measures how good or bad a system is. In image classification, well-known networks only consider modifying the network structures and applying cross-entropy loss at the end of the network. However, using only cross-entropy loss causes a network to stop updating weights when all training images are correctly classified. This is the problem of the early saturation. This paper proposes a novel cost function, called mixture separability loss (MSL), which updates the weights of the network even when most of the training images are accurately predicted. MSL consists of between-class and within-class loss. Between-class loss maximizes the differences between inter-class images, whereas within-class loss minimizes the similarities between intra-class images. We designed the proposed loss function to attach to different convolutional layers in the network in order to utilize intermediate feature maps. Experiments show that a network with MSL deepens the learning process and obtains promising results with some public datasets, such as Street View House Number (SVHN), Canadian Institute for Advanced Research (CIFAR), and our self-collected Inha Computer Vision Lab (ICVL) gender dataset.

## 1. Introduction

Many challenging tasks in computer vision, speech recognition, and natural language processing are quickly resolved thanks to the advances in deep convolutional neural networks. This success is attributed to the automatic selection of features to replace heuristic feature selection, because feature selection is the most crucial step in any artificial intelligence system. In particular, in image classification, instead of manually designing features for a system with prior information and expert knowledge, a deep neural network provides a complete procedure for both feature extraction and image classification.

To obtain features at a higher level of abstraction, deep network structures have been mostly changing from shallow AlexNet [1], the Visual Geometry Group network (VGG) [2], and inception networks [3-7] to very deep residual networks [8-10] and their derivatives [11, 12], where accuracy surpasses the human level. In addition to the network structure, several techniques have been proposed, such as careful initialization [7, 13], the activation function [14, 15], various types of pooling [16], strategies for regularization [17, 18], and numerous loss functions [19-25], in order to deal with problems of gradient vanishing and overfitting.

Although many loss functions were introduced in the literature, only some are used in deep neural networks for training a classification model. Most of the well-known network structures [1, 2, 6, 9] use only a single loss function (cross-entropy loss). As defined in information theory [26], cross-entropy is calculated as the homogeneity between ground truth and the predicted probability densities of an instance. For a long time, cross-entropy loss showed effectiveness in the training of a deep model. However, when all training samples are correctly classified (two densities are identical), the training error is nearly zero, thus indicating the weights updated are too small, as the training loss almost always becomes a constant value. Thus, the training process leads to a converged state and terminates. However, testing accuracy is not as high as expected. We describe this phenomenon as early saturation. Moreover, in the conventional methods, the loss function obtains the feature maps only from the last convolutional layer to perform the final classification step.

In this paper, we propose a novel cost function, called mixture separability loss (MSL), consisting of between-class loss and within-class loss. Between-class loss works like the traditional loss, maximizing the differences in images from multiple classes, whereas within-class loss is designed to minimize the similarities of images from a single class. In addition, we claim that the loss function attached to the preceding layers provides valuable information that the last layer cannot provide. The major contributions of this paper are threefold:

- We propose a novel cost function (mixture separability loss) to simultaneously maximize the differences and minimize the similarities between images in the training phase.
- We redesign the network to attach MSL in several positions of the network to make it learn not only the feature map of the last layer but also the features in the earlier layers as well.
- We design a novel mixture separability module, including between-class losses and within-class losses, which is flexible and can easily be incorporated into other available deep neural networks.

The remainder of the paper is organized as follows. Section 2 discusses the related works. Section 3 introduces the new loss function and the module to plug into different positions of convolutional networks. Section 4 demonstrates the results on public datasets, such as Street View House Number (SVHN), Canadian Institute for Advanced Research (CIFAR), and our self-collected Inha Computer Vision Lab (ICVL) gender dataset and discusses details of a network analysis. Finally, Section 5 provides the conclusions and future works.

## 2. Related works

Since the onset of deep learning with AlexNet [1] to the state-of-the-art deep residual networks [8, 10, 12], only single loss (cross-entropy loss) has been applied to train a deep model. Cross-entropy loss, or the softmax loss, is the combination of multinomial logistic and softmax functions. It changes predictions to non-negative values and normalizes them to obtain the probability distribution over classes. Cross-entropy loss has been widely applied in image classification because of its simplicity and effectiveness. To avoid the overfitting problem from using the conventional softmax loss, Liu et al. [21] introduced a large-margin softmax (L-softmax) loss that creates an angular margin to the angle between the input feature vector and the column of the weight matrix. The conventional softmax loss can be considered a special case of L-softmax when the margin among classes is set to 1.

Other researchers [19, 20] attempted to replace cross-entropy loss with a squared-hinge loss (the support vector machine loss). Their experiments, using a shallow network on some public datasets, show that the network trained using the SVM loss obtains better results than using the softmax loss. Shaham et al. [23] used a contrastive loss with a Siamese network [24] to learn the similarity between a pair of data instances. First, a pair of images is fed through two identical deep neural networks to obtain feature vectors that are used to compute the Euclidean distance. The contrastive loss is accumulated based on either matching-pair or non–matching-pair distances. Schroff et al. [25] considered three instances per loss function, namely, triplet loss instances (anchor, positive, and negative). The triplet loss minimizes the distance between anchor and positive instances; meanwhile, it simultaneously maximizes the distance between anchor and negative instances.

Similar to our work, Xu et al. [22] proposed a method to incorporate multiple losses, including softmax loss, pairwise ranking loss, and LambdaRank loss, for the image classification problem. These losses come from different theoretical motivations. Pairwise ranking loss assigns a class label from two classes of an image. The LambdaRank loss calculates the desired gradient directly, rather than computing it from a loss function. These losses are used with backpropagation to train a deep neural network. In the testing phase, multiple losses are fused with average pooling to produce the final prediction.

## 3. Proposed model

In this section, we introduce the concepts of global average pooling [6] and the MSL loss function, including between-class and within-class losses used in the training process. Then, we integrate these concepts to form a mixture separability module (MSM). Finally, we construct the mixture separability network (MSN) by injecting the MSM into the convolutional layers of well-known networks.

### 3.1. Global average pooling

As shown in the network structure of old versions, such as AlexNet [1] and VGG [2], many parameters cause the training and testing process to slow down, and they are prone to overfitting. More than 80% of the parameters are obtained from the fully connected (FC) layer, the last layer of the network. Therefore, to reduce the number of parameters and avoid overfitting, we use the global average pooling operation [6, 27].

We assume that the dimensions of the feature map after a convolutional layer for a single image have the size $m \times m \times d$, where $m$ is both width and height, and $d$ is the depth of the feature map. The global average pooling layer calculates the average value of an $m \times m$ single feature map over $d$ dimensions to obtain a vector, $[1 \times d]^T$. The obtained values from the vector are then the weighted summation in $c$ different ways (the FC layer), where $c$ is the number of classes, to obtain a class score of size $1 \times c$ for each image. This significantly reduces the number of parameters in the FC layer of conventional networks, while classification performance is still maintained [27].

### 3.2. Between-class loss

The loss function is the backbone in building an effective artificial intelligence system, because it measures how good or bad a system is. In image classification, a well-known function measuring classification performance is the cross-entropy loss function, or log loss function. During the training process, this loss function penalizes wrongly classified images. In other words, minimizing the cross-entropy losses means maximizing the difference in images between classes.

We assumed $N$ training images, $X = (x^{(1)}, \dots, x^{(N)})$, with corresponding labels, $Y = (y^{(1)}, \dots, y^{(N)})$, in a mini-batch. We fed the obtained convolutional layer feature maps through average pooling and the FC layer, as described in Subsection 3.1, to obtain a three-dimensional matrix of size $N \times 1 \times c$, where $c$ is the number of classes. Each image in the mini-batch produces $q^{(i)} = [1 \times c]^T$ vectors, denoted as $q^{(i)} = [q_1^{(i)}, q_2^{(i)}, \dots, q_c^{(i)}]^T$ where $i = 1..N$. Equation (1) defines cross-entropy [26], where $\mathbb{1}[y^{(i)} = j]$ is equal to 1 when label index $y^{(i)}$ of image $x^{(i)}$ is $j$; otherwise, it equals zero, and $p_j^{(i)}$ is the estimated probability density of image $x^{(i)}$ belonging to class $j$:

$$\mathcal{L}_B = -\frac{1}{N} \sum_{i=1}^{N} \sum_{j=1}^{c} \mathbb{1}[y^{(i)} = j] \, log(p_j^{(i)}) \qquad (1)$$

$$p_j^{(i)} = \frac{exp(q_j^{(i)})}{\sum_{k=1}^{c} exp(q_k^{(i)})} \qquad (2)$$

During the training process, minimizing the cross-entropy losses led to maximizing the differences in images between classes in a feature space. Therefore, we termed cross-entropy loss the between-class loss of images in different classes. This loss is the conventional loss embedded in most state-of-the-art networks, such as VGG [2], GoogleNet [6], the residual network (ResNet) [10], and the wide residual network (WRN) [9] for image classification.

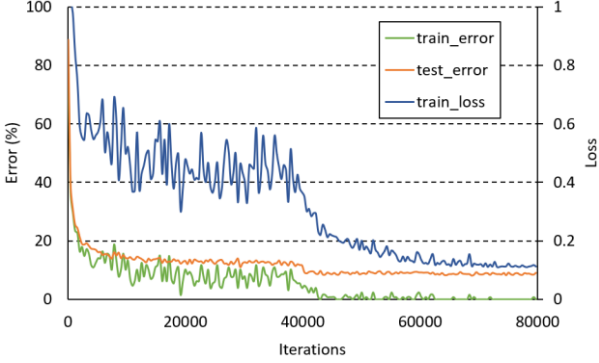

***Fig. 1.*** *Training/testing error and loss on the CIFAR-10 dataset of ResNet20 using a cross-entropy loss function*

As shown in Figure 1, the between-class loss gets smaller as more images are correctly classified during training. The training error (green line) and testing error (orange line) are referenced using the left axis, whereas the training loss (blue line) is referenced using the right axis. From 60,000 iterations, the training error is nearly zero, and the testing error cannot be improved. In the training process, by minimizing the between-class losses, we updated the weights so that the differences in images from various classes increased. This process can be called class discretization to differentiate images in the classes. Moreover, when the probability density of an image and its ground-truth distribution are nearly identical (very high confidence), the between-class loss is not reduced. Thus, the weights are no longer updated, and the training process is terminated because nothing more remains to be learned. The accuracy with the training data was very high, but the accuracy with the testing data was not as high as expected. This is the problem of early saturation.

### 3.3. Within-class losses

We propose a method to measure the similarities of images in a single class to make networks learn a generalization of the class. Then, we incorporate the similarities and differences to ensure the network learns both the generalization and the discretization simultaneously.

Among $N$ images of a mini-batch, we denote the number of images from class $j$ as $\mu_j$, an in-class distance of images $d_j$ in class $j$ is defined in Equation (3), where $q^{(i)}$ is the feature vector of image $i$ after the FC layer.

$$d_j = \frac{1}{\lambda} \sum_{i,i^*}^{\mu_j} \sum_{k=1}^{c} \left\| q_k^{(i)} - q_k^{(i^*)} \right\|_2 \qquad (3)$$

$$\lambda = \frac{\mu_j!}{2!\,(\mu_j - 2)!} \qquad (4)$$

In Equation (3), in-class distance $d_j$ is calculated as the Euclidean distance between the non-overlapped pairs of images of class $j$ in the feature space, divided by the number of pairwise combinations, $\lambda$, where $\lambda$ represents the number from selecting a pair of images, $x^{(i)}$ and $x^{(i^*)}$, from class $j$ as shown in Equation (4).

We define the within-class loss by determining if in-class distance $d_j$ is larger than an adaptive threshold value, $\xi$, as seen in Equation (5). When the loss is nearly unchanged after a number of iterations (100 in this work), we designed $\xi$ to decrease by 10% to finely update smaller changes of the in-class distance in order to deepen the learning process. In this work, we used squared-hinge loss [20, 28] for measuring the similarities of images from a class, defined in Equation (5):

$$\mathcal{L}_W = \sum_{j=1}^{c} \max\bigl(0, d_j - \xi\bigr)^2 \qquad (5)$$

In the training process, we update the weights by using both between-class and within-class losses, because we calculate the gradient in the entire mini-batch, as shown in Equation (6):

$$\mathcal{L}(\theta) = \mathcal{L}_B\bigl(\theta;\, x^{(i)}, y^{(i)}\bigr) + \mathcal{L}_W\bigl(\theta;\, x^{(i)}, y^{(i)}\bigr) \qquad (6)$$

With a larger mini-batch size, a better model is obtained, because with the larger mini-batch size, the gradient of the loss firmly points to the global minimal value, rather than the local one, in the optimization process. The MSL becomes the cross-entropy loss when only a single image per cluster ($\mu_j = 1$) is fed into the network during training time.

### 3.4. Mixture separability network

In conventional networks, the number of convolutional layers is increased (from dozens to hundreds) in order to learn high-level image representation [10]. They only use feature maps from the last layer to calculate the classification loss of the network. However, the losses can be calculated from different convolutional layers of the network. Those losses are then used to form the final loss for optimization by taking the average. By doing so, the feature maps from earlier layers are forced to directly contribute to the weight update process.

In this work, we use state-of-the-art networks, including VGG [2], ResNet [8, 10] and WRN [9], as baselines (see Table 1) to construct the MSN. Baselines are the networks where cross-entropy loss is calculated in the last layer. For comparison with other deep network structures that consist of only three max-pooling layers [8, 9], we used a modified version of VGG16 [2]. In structures of the modified version of VGG16, we processed three $3 \times 3$ convolutional layers in the first convolutional blocks (conv_1) and four $3 \times 3$ convolutional layers in the last three convolutional blocks (conv_2, conv_3, conv_4). In the residual-network family [8, 10] including ResNet3, ResNet9, ResNet18, ResNet25, with $K = 3, 9, 18, 25$, respectively, we invoke batch normalization [7] and the rectified linear unit (ReLU) [14] before each convolutional layer [8]. To form the convolutional block in the residual network, we sequentially stacked $K$ residual blocks on top of each other. Finally, the structure of WRN28/10 is similar to the residual network structure, except for the number of convolutional layers and a widening factor in the residual blocks. The single convolutional block is a set of convolutional layers before the max pooling operation. We obtain the final classification results by using global average pooling followed by the cross-entropy loss function. It is highly recommended to refer to other papers [2, 8-10] to understand the baseline structures.

**Table 1** Details structure of the baseline networks including VGG16, ResNetK (*K*=3, 9, 18, 25), and WRN28/10

|        | VGG16 | ResNetK | WRN28/10 |
|--------|-------|---------|----------|
| conv_1 | $[3 \times 3\ conv\ 64] \times 3$ | $3 \times 3\ conv\ 16$ | $3 \times 3\ conv\ 16$ |
|        | max pooling | max pooling | max pooling |
| conv_2 | $[3 \times 3\ conv\ 128] \times 4$ | $\begin{bmatrix}3 \times 3\ conv\ 16\\3 \times 3\ conv\ 16\end{bmatrix} \times K$ | $\begin{bmatrix}3 \times 3\ conv\ 16 \times 10\\3 \times 3\ conv\ 16 \times 10\\3 \times 3\ conv\ 16 \times 10\end{bmatrix} \times 3$ |
|        | max pooling | max pooling | max pooling |
| conv_3 | $[3 \times 3\ conv\ 256] \times 4$ | $\begin{bmatrix}3 \times 3\ conv\ 32\\3 \times 3\ conv\ 32\end{bmatrix} \times K$ | $\begin{bmatrix}3 \times 3\ conv\ 32 \times 10\\3 \times 3\ conv\ 32 \times 10\\3 \times 3\ conv\ 32 \times 10\end{bmatrix} \times 3$ |
|        | max pooling | max pooling | max pooling |
| conv_4 | $[3 \times 3\ conv\ 512] \times 4$ | $\begin{bmatrix}3 \times 3\ conv\ 64\\3 \times 3\ conv\ 64\end{bmatrix} \times K$ | $\begin{bmatrix}3 \times 3\ conv\ 64 \times 10\\3 \times 3\ conv\ 64 \times 10\\3 \times 3\ conv\ 64 \times 10\end{bmatrix} \times 3$ |

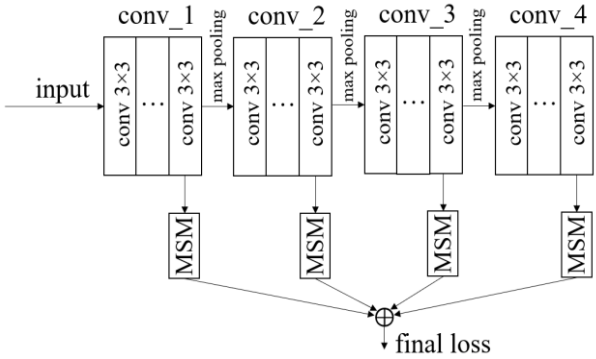

*Fig. 2. Overall flowchart of the proposed network*

To construct the MSN, we first designed a mixture separability module (MSM) that consists of the global average pooling layer, the FC layer, and the MSL. We used the MSM to replace the FC layer in the conventional deep neural network for the loss calculation. The novel loss provided by the MSM is different from the one used in the conventional neural network, because the proposed loss includes both between-class and within-class losses, as mentioned in Subsections 3.2 and 3.3.

We constructed the corresponding MSN versions by attaching the MSM to the last convolutional layer in each convolutional block of the baselines. At each insertion point, we concurrently compute the between-class and within-class losses. The final loss of the MSN is the average of all losses previously calculated. Figure 2 shows the overall flowchart of the proposed MSN. Experiments in Section 5 show that the classification performance produced by MSN on several public datasets is superior to that produced by the baselines.

## 4. Experimental results

We used the following environments and tools for this study: a PC with two GeForce GTX 1080 graphics cards and 32 GB of RAM running Ubuntu 14.04 LTS with an installed TensorFlow tool [29]. We set the mini-batch size to 128 at a fixed momentum of 0.9. The total number of iterations was 80,000, and we set the initial value of the learning rate to 0.01, which kept reducing by 10% every 20,000 iterations. The initial value of marginal distance $\xi$, see Equation (5), was set to 0.5 and was reduced by 10% when the within-class loss did not change after 100 iterations. For comparison with other baseline networks, we followed Goodfellow et al. [15] for the data pre-processing, training, and testing hyper-parameters.

### 4.1. CIFAR-10 dataset

CIFAR-10 [30] consists of 60,000 $32 \times 32$ RGB images from 10 classes with 6000 images per class. From the dataset, 50,000 and 10,000 images were used for training and testing, respectively. We performed the experiments by alternately applying cross-entropy loss or the MSL with VGG16, ResNet3, ResNet9, ResNet18, ResNet25, and WRN28/10. For data augmentation, we applied global contrast normalization, a whitening process, and horizontal flipping on the input images.

As shown in Figure 3, the testing error was reduced significantly as the network got deeper and wider, from network in network (NIN) (8.81%) to ResNet25 (5.97%) and WRN28/10 (4%). The results from NIN are taken from [22] in which multiple loss was used instead of MSL. For other deep networks, such as VGG16 and ResNet3, the testing errors were further reduced when the new cost function, MSL, was applied. Classification errors with VGG16 and 20-layer ResNet3 networks were reduced to 7.11% from 7.55%, and to 7.07% from 7.45%, respectively. More interestingly, with the deeper residual-network family, 56-layer ResNet9 obtained a 6.97% testing error using cross-entropy loss, and 6.25% using MSL. Adding more layer versions of the 110-layer ResNet18 reached 6.43%, which is not better than the 56-layer ResNet9 with MSL. In our experiment, the deepest version of ResNet was 152-layer ResNet25, which reached 5.97% with cross-entropy loss and was further reduced to 5.48% with MSL. We confirmed the effectiveness of MSL with the WRN28/10 wider network, which went from 4.00% to 3.72% (a new state-of-the-art rate for the CIFAR-10 dataset).

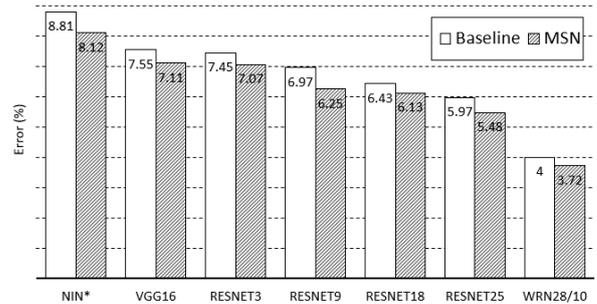

*Fig. 3. Comparison of classification error on the CIFAR-10 test set for baseline deep networks and its corresponding MSN versions*

### 4.2. CIFAR-100 dataset

CIFAR-100 [30] consists of 60,000 32 × 32 RGB images from 100 classes with only 600 images per class. The data were split into 50,000 and 10,000 images for training and testing, respectively. The above deep networks were used to confirm the performance of the new cost function. Like the process with CIFAR-10, we applied global contrast, a whitening process, and horizontal flipping on the training input images. As there are 100 classes in this dataset, we set the mini-batch size to 512 instead of 128 to ensure at least several images from a single class were fed into the network during training. It is important, because the within-class loss has more effect on a group (cluster) of images than a single image from a class.

As shown in Figure 4, the results of the NIN network from [22] were used, in which multiple losses are used instead of MSL. CIFAR-100 has more classes to categorize, but fewer images to learn from in a class. This results in a higher testing error, but the model keeps getting better when the networks get deeper. NIN [22] produced the highest testing error, at 33.53% and 31.47% with multiple loss functions. The networks with MSL generated better results than those with single cross-entropy only; for example, testing error with VGG16 was reduced to 30.53% from 31.85%, and with the 20-layer ResNet3, the error was reduced to 30.28% from 31.61%. The results with VGG16 and ResNet3 were similar, perhaps because they have a similar number of convolutional layers. The deeper residual networks added more convolutional layers to residual blocks and yielded better results (from 27.01% to 26.80% with ResNet9 and ResNet18). The 56-layer ResNet9 version with MSL performed better than the one with 110-layer ResNet18 and conventional cross-entropy, which again confirms the effectiveness of the new loss function. It indicates that the testing error can be reduced either by adding more convolutional layers to residual blocks or by using the proposed loss function. Finally, the WRN28/10 network (no dropout) with MSL produced a testing error of 18.29%, which is the best up to now when using CIFAR-100.

### 4.3. SVHN dataset

SVHN [31] is a real-world image dataset, obtained from house numbers in Google Street View images. It consists of 630,420 32 × 32 RGB images, of which 73,257 were used for training, 26,032 images for testing, and the other 531,131 images were used for extra training. Similar to the CIFAR dataset, we performed global contrast normalization, a whitening process, and horizontal flipping on the input images for data augmentation.

As shown in Figure 5, we conducted the experiments on the SVHN dataset with several deep neural networks: NIN, VGG16, ResNet3, ResNet9, ResNet18, ResNet25, and WRN28/10. We used the results with NIN from [22], whereas those from other networks were obtained by training the networks from scratch. We confirmed the effectiveness of the new cost function on the SVHN public dataset. The testing error was reduced by at least 0.3% on all networks when the MSL was used instead of cross-entropy loss. The experiments show that the VGG16 network with MSL (1.92%) performed better than the traditional ResNet3 (2.25%), ResNet9 (2.21%), ResNet18 (2.18%), and ResNet25 (2.14%) with cross-entropy. A possible reason could be that a very deep network is too large for this relatively simple dataset. The best classification performance was obtained by using WRN28/10 and its corresponding MSN version, which reduced the classification error to 1.64% from 1.81%. A deepening network with more convolutional layers and a widening network with more constraint on the loss are possible methods to improve classification performance.

### 4.4. ICVL gender dataset

The next experiment was testing recent networks and the new cost function to solve a real-world problem: gender classification from a distance in a surveillance system. We collected data from our surveillance system. There were 10 cameras installed on the ceiling or on a lamp column to capture both indoor and outdoor scenes. First, we manually cropped human bounding boxes from full images. Then, we resized these boxes to a 144-pixel height to preserve the human ratio. To make the gender data compatible with a deep learning method, we padded zero values to the left and right sides of the resized images to obtain a 144-pixel width. Our gender dataset consists of 13,000 color images at $144 \times 144$ in the two classes (male and female). Each gender class had an equal number of images (6,500). The number of training images and testing images were 11,000 and 2,000, respectively. To train the model, we performed data augmentation, including global contrast normalization, the whitening process, and horizontal flipping, as in the previous experiments. Figure 6 shows examples of the ICVL gender dataset.

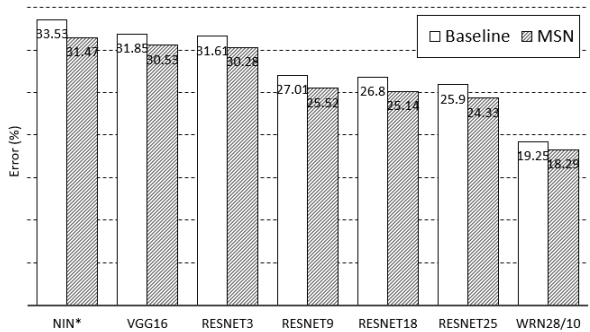

***Fig. 4.*** *Comparison of classification error on the CIFAR-100 test set for baseline deep networks and the corresponding MSN versions*

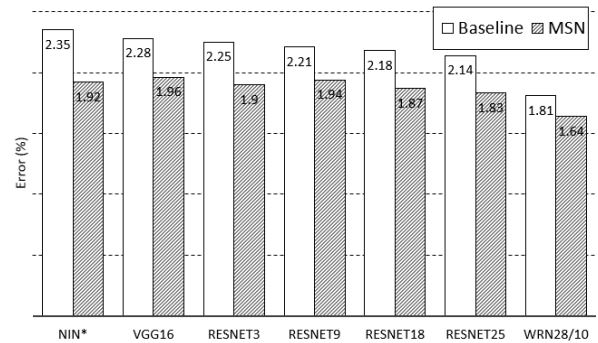

***Fig. 5.*** *Comparison of classification error on the SVHN test set for baseline deep networks and the corresponding MSN versions*

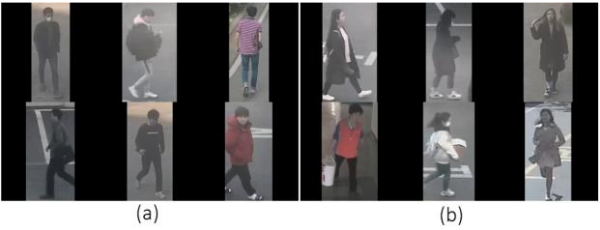

***Fig. 6.*** *ICVL gender dataset, (a) male, (b) female*

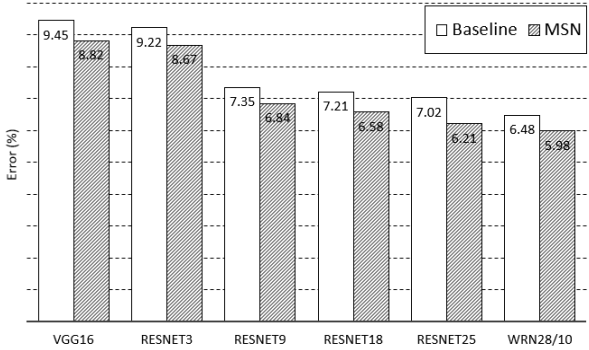

***Fig. 7.*** *Comparison of classification error on the ICVL gender test set for baseline deep networks and the corresponding MSN versions*

As shown in Figure 7, using the new MSL loss function to build the deep convolutional model is recommended. It yielded a nearly 0.5% lower testing error than a corresponding model using a conventional cross-entropy loss function and reduced the error to 8.82% from 9.45% with VGG16. The deeper network, ResNet25 and its corresponding MSN version, reduced the testing error even more, to 6.21% from 7.02%. The deeper and wider WRN28/10 produced the best result at 6.48%, which was further improved to 5.98% by using MSL again. The experiments proved that simultaneously learning the differences between classes and the similarities within a class avoided the early saturation problem, and thus provides better accuracy.

### 4.5. Network structure analysis

We provide further analysis for VGG16 with the CIFAR-100 dataset to demonstrate the influence of adding the MSM and the batch size to the final testing results. The reason for selecting VGG16 and CIFAR-100 are: i) the VGG16 network is relatively easy to set up in various configurations, and ii) the CIFAR-100 dataset with 100 classes is relatively large and difficult to learn.

To make the deep network learn from the early convolutional layers, we added several MSMs to the network at the last convolutional layer in each convolutional block (see Subsection 3.4 for more details). There were four such modules. We obtained the final loss by calculating the loss average from all added modules. We set all hyperparameters as in the experiment with CIFAR-100 (see Subsection 4.2).

**Table 2** Classification error on the CIFAR-100 test set of VGG16 in different configurations, (I, II, III, IV) adding MSM to the last convolutional layer of the 1st, 2nd, 3rd, and 4th convolutional blocks.

|  | I | II | III | IV | Classification error (%) |
|---|---|---|---|---|---|
| Config1 |  |  |  | ✓ | 31.25 |
| Config2 |  |  | ✓ | ✓ | 30.98 |
| Config3 |  | ✓ |  | ✓ | 31.04 |
| Config4 | ✓ |  |  | ✓ | 31.08 |
| Config5 |  | ✓ | ✓ | ✓ | 30.61 |
| Config6 | ✓ |  | ✓ | ✓ | 30.65 |
| Config7 | ✓ | ✓ | ✓ | ✓ | 30.53 |

As shown in Table 2, the testing error from VGG16 on CIFAR-100 is 31.25% when MSM is added to the fourth convolutional block (config1). The result improves when more MSMs are attached to the preceding convolutional blocks. Classification performance under config2, config3, and config4 were 30.98%, 31.04%, and 31.08%, respectively. The performance is similar because the final loss is calculated from two convolutional blocks. Using the loss from two deeper convolutional layers, the result with config2 is slightly better than that with config3 and config4. The classification error is further reduced when the loss is accumulated from three previous layers, as in config5 and config6 (30.61% and 30.65%, respectively). Finally, the best performance was obtained by adding an MSM to all convolutional blocks (config7) in VGG16. In all configurations, the MSM in the last convolutional block (conv_4) is necessary because feature maps from a high level of abstraction are very important. Excluding the MSM from the last convolutional block (conv_4) reduces the depth of the network, thus resulting in worse classification performance. The results suggest that classification performance can be improved by forcing the network to learn from feature maps in the preceding layers, rather than the last layer only.

We investigated the influence of mini-batch size on the testing results when using the new loss function and compared it with the one using cross-entropy loss. We attached all losses at the last convolutional layer of the VGG16 network. The hyper-parameters were set the same as in the experiments with CIFAR-100 (Subsection 4.2) except for the batch size, which was tuned with values of 64, 128, 256, and 512. Table 3 shows the testing accuracies.

**Table 3** Comparison of classification errors on the CIFAR-100 test set of VGG16 using different batch sizes

|  | Classification error (%) | |
|---|---|---|
|  | Cross-entropy | MSL |
| batch size 64 | 31.82 | 31.83 |
| batch size 128 | 31.81 | 31.77 |
| batch size 256 | 31.84 | 31.06 |
| batch size 512 | 31.85 | 30.53 |

As shown in Table 3, the testing errors using cross-entropy are slightly different with different numbers of batch sizes (31.82% and 31.85% with a batch sizes of 64 and 512, respectively). The differences are small because the cross-entropy function calculates the average loss in a batch. It is almost the same, regardless of whether the batch sizes are low or high. The calculation changed when the MSL was used to train the network. We obtained testing errors of 31.81% and 30.53% with batch sizes of 64 and 512, respectively, which is a significant improvement. The reason is in the method MSL uses to calculate the loss. In the MSL, except for cross-entropy loss, we calculate the similarities between a pair of images. When the batch size was small (64, in this case), the in-class distance loss did not have any effect because only a single image in a given class was fed into the training. By contrast, when the batch size was 512, several images from a class were fed into the training (at least five in this case). MSL is designed to maximize differences between the images in classes and minimize similarities of the images in a class, giving it an advantage. Thus, the testing error was reduced.

Finally, we selected the best configuration of the MSN (config7) to show the effectiveness of the MSL compared to the original VGG16 with cross-entropy in terms of error and loss. We set the mini-batch size to 512 for the comparison. All hyper-parameters were set according to Subsection 4.2. The behavior of training errors and losses are summarized in Figure 8 where errors and losses are referenced by using the left and right axes, respectively. As shown in Figure 8, the training error (the blue line) produced by using MSN converges faster than that provided by VGG16 (the red line). In addition, when the VGG16–MSN training error approaches zero (from 60,000 iterations), its training loss (the green line) gradually reduces, whereas the training loss with VGG16 (the orange line) is the same. The best model produced by VGG16–MSN and VGG16 were obtained with iterations of 76,532 and 41,145, respectively. This indicates that the VGG16–MSN model still learns from the examples, even though the training error is very low. Hence, a lower classification error in the CIFAR-100 testing data is obtained, from 31.85% with VGG16 to 30.53% with VGG16–MSN, as shown in Figure 9.

## 5. Conclusions

In this paper, we proposed a novel loss function, MSL, that includes between-class and within-class loss. Between-class loss maximizes the differences between images from different classes, whereas within-class loss minimizes the similarities between images in a single class. Using the proposed losses, a deep model further learns to avoid the early saturation problem, a common phenomenon in deep neural networks. The proposed loss function is flexible, so it can be plugged into any position of the network. By adding the loss function at different convolutional layers, we force the network to learn from feature maps in early layers, rather than the last layer only, as in conventional methods. The experiments showed that the classification performance of the proposed network achieved competitive performance on some public datasets, compared to conventional networks. It is worth mentioning that the classification accuracy can be further improved by applying the MSL to more advanced pyramidal residual networks [32, 33] which are currently state of the art on CIFAR-10 (2.96%) and CIFAR-100 (16.4%) datasets.

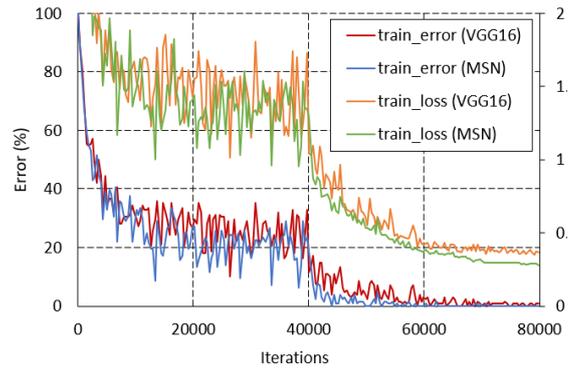

*Fig. 8.* Training error/loss on CIFAR-100 with VGG16 and the corresponding MSN version

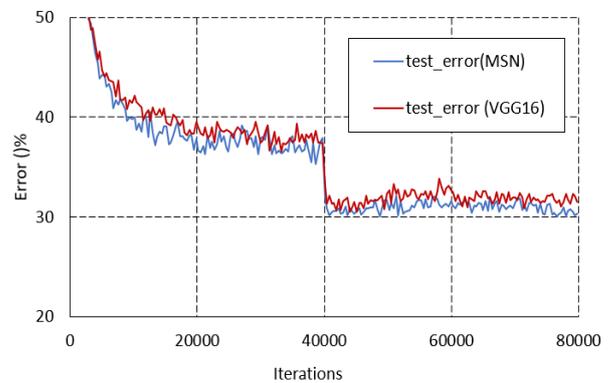

*Fig. 9.* Testing error on CIFAR-100 with VGG16 and the corresponding MSN version

## 6. Acknowledgments

This work was supported by the Industrial Technology Innovation Program, "10052982, Development of multi-angle front camera system for intersection AEB," funded by the Ministry of Trade, Industry, & Energy (MOTIE, Korea).